%% file: bwb_icip2021.tex
\documentclass{article}

\usepackage{amsmath,graphicx}
\usepackage[preprint]{spconf}
\usepackage{graphics}
\usepackage{tikz}
\usepackage{subfig}
\usepackage{multirow}
\usepackage{hyperref}
\usepackage{todonotes}
\usepackage{amssymb}

\include{defines}

\newcommand{\rotcell}[1]{%
  \rotatebox[origin=c]{90}{ #1 }%
}

\newcommand\copyrighttext{%
  \footnotesize \copyright\ IEEE 2021. Personal use of this material is permitted. Permission from IEEE must be obtained for all other uses, in any current or future media, including reprinting/republishing this material for advertising or promotional purposes, creating new collective works, for resale or redistribution to servers or lists, or reuse of any copyrighted component of this work in other works.}
\newcommand\mycopyrightnotice{%
\begin{tikzpicture}[remember picture,overlay]
\node[anchor=north,yshift=-10pt,outer sep=0pt] at (current page.north) {\fbox{\parbox{\dimexpr\textwidth-\fboxsep-\fboxrule\relax}{\copyrighttext}}};
\end{tikzpicture}%
}

\title{\vspace{-0.3cm}Curiously effective features for image quality prediction}
\name{%
  \parbox{\linewidth}{\centering
    S\"oren Becker$^1$,
    Thomas Wiegand$^{1,2}$, and
    Sebastian Bosse$^1$
  }%
}%

\address{$^1$Fraunhofer Heinrich Hertz Institute, Berlin, Germany \quad
$^2$Technical University of Berlin, Germany}

\copyrightnotice{\copyright\ IEEE 2021}
                 \toappear{To appear in {\it Proc.\ ICIP 2021,
                 September 19-22, 2021, Anchorage, Alaska, USA}}

\begin{document}
%
\maketitle
\mycopyrightnotice

\begin{abstract}
The performance of visual quality prediction models is commonly assumed to be closely tied to their ability to capture perceptually relevant image aspects. Models are thus either based
on sophisticated feature extractors carefully designed from extensive domain knowledge or optimized through feature learning. In contrast to this, we find feature extractors constructed from random noise to be sufficient to learn a linear regression model whose quality predictions reach high correlations with human visual quality ratings, on par with a model with learned features. We analyze this curious result and show that besides the quality of feature extractors also their quantity plays a crucial role - with top performances only being achieved in highly overparameterized models.
\end{abstract}

\section{Introduction}
\label{sec:intro}

Accurate and reliable prediction of visual quality  is  crucial for a wide
range of applications in image processing, computer graphics and multimedia
communication including compression, enhancement and restoration \cite{chandler2013seven}.
However, while humans judge visual quality seamlessly, the computational prediction of human quality ratings, known as mean opinion scores (MOS), still remains challenging.
Computational approaches to quality estimation are typically distinguished by
the availability of the reference image into full reference (FR), reduced
reference (RR) and no reference (NR) models. 
The reference availability of a
quality model relates to its scope of application as e.g. in a compression
scenario the reference signal is generally available \cite{Bosse2019b}, while in superresolution
or enhancement tasks it might not even exist \cite{chandler2013seven}.\\
The typical design principle of quality models consists of different stages,
including feature extraction, feature pooling, and a regression that maps the
(pooled) features to a scalar quality estimate.  Its technological
development follows, in general lines, the
development in other fields in computer vision, leading from approaches
centered around feature engineering  over feature learning-based methods 
to today's end-to-end learned models. 
Classical models for perceptual quality prediction are based on sophisticated,
handcrafted features that either imitate specific
characteristics of the human visual system or capture the statistics of natural
images and their perceptually relevant deviations. One of the most prominent and
seminal examples for scene statistic-based FR quality estimation is SSIM
\cite{wang2004image} that inspired many other models \cite{Wang2003, zhang2011fsim}. 
Feature engineering approaches led to a remarkable success, especially in
the domain of FR quality estimation. However, it is often speculated that
the underlying domain knowledge is too limited or incomplete to
capture the full complexity of human visual (quality) perception.  This led to quality
models leveraging methods of feature learning  that allowed
advances in RR and NR quality estimation. Some models borrow ideas from
both feature engineering and feature learning, as quality features correspond to 
\textit{learned} parameters of the (assumed) probability function of certain
\textit{engineered} features
\cite{Mittal2012, Saad2012}. Models qualifying for an unmistakable definition of feature learning
employ dictionary learning \cite{Ye2012a, Zhang2015}, principal component analysis
\cite{SPCA} or neural networks \cite{Zhang2018, Bhardwaj2020}. Here, features are learned in an
unsupervised \cite{Ye2012a, Zhang2015, SPCA} or self-supervised \cite{Bhardwaj2020} fashion
or based on a task different to quality estimation \cite{Zhang2018, chetouani2020image}.
With the success of deep learning in other fields of computer
vision, researchers started to devise end-to-end optimized neural network-based
approaches to FR, RR and NR quality estimation \cite{Bosse2018, ding2020image, Prashnani2018, Bosse2018distsens}. Although the end-to-end design
blurs the lines between feature extraction, pooling and
regression, early to middle layers, foremost
convolutional layers, can be regarded as implementing feature extractors, while later layers,
mostly fully connected layers, can be considered to implement the regression.\\
Regardless of whether features are handcrafted from domain knowledge or learned by the model, a common underlying assumption appears to be that plausible features for image quality prediction relate to properties of human visual perception or natural scene statistics. 
In this paper, we reevaluate this hypothesis. 
To our surprise, we find that 
even feature extractors constructed from pure noise suffice to optimize a linear regression that 
achieves high correlation with human quality ratings, similar to a regression 
from learned features. In subsequent analyses we relate this result to the lottery ticket hypothesis \cite{Ramanujan_2020_CVPR} and the double descent effect \cite{belkin2019reconciling} that have recently been described in the machine learning literature.
The computational framework for our study is introduced in \Secref{sec:computational_framework} and details of the experimental setup are described in \Secref{sec:experimental_setup}. Results and subsequent analyses are reported in \Secref{sec:results} and our conclusions are derived in Sec.~\ref{sec:conclusion}.

\section{Computational framework}
\label{sec:computational_framework}
The computational framework for our evaluation is inspired by CORNIA \cite{Ye2012a}, one of the earliest visual quality models based on feature learning. CORNIA offers an attractive basis for our experiments as on the one hand it achieves high correlations with human quality ratings but on the other hand follows an elegant yet comparatively simple architecture. In particular, feature extractors and regression model are decoupled and optimized successively, allowing us to inject and evaluate different feature extractors irrespective of the rest of the model.\\
For a given image $I$, CORNIA starts by extracting a set of descriptors $\mathbf{X}_{d \times n}=[\mathbf{x}_1, ..., \mathbf{x}_n]$ with each descriptor $\mathbf{x}_i \in \mathbb{R}^d, d=h \cdot w$, corresponding to a randomly sampled local image patch of h$\times$w pixels. Each descriptor is standardized and the set of descriptors is whitened by zero component analysis (ZCA). Preprocessed descriptors from a set of training images are clustered into $k$ centroids via k-means. All centroids are normalized to unit length and stored in a matrix $\mathbf{D}_{d \times k}$ which represents a visual codebook.
The codebook is used to extract features from a given image. Specifically, local descriptors $\mathbf{X'}_{d \times l}=[\mathbf{x'}_1, ..., \mathbf{x'}_l]$ of an image are encoded as $\mathbf{S}_{k \times l} = \mathbf{D}^T \mathbf{X'}$. 
In addition, CORNIA applies a non-linear soft-encoding function to further process the extracted features. Let $s_{ij}$ denote the dot product between the $i$-th centroid and the $j$-th descriptor. The $j$-th soft-encoded descriptor is then given by 
\begin{align*}
\mathbf{c}_j = [&\text{max}(s_{1, j}, 0), ..., \text{max}(s_{k, j}, 0), \\
       &\text{max}(-s_{1, j}, 0), ..., \text{max}(-s_{k, j}, 0)]^T.
\end{align*}
The soft-encoding step results in a matrix $\mathbf{C}_{2k \times l}$ which is reduced to a feature vector $\boldsymbol{\beta} \in \mathbb{R}^{2k}$ in the final step of feature extraction, with $\beta_j = \max \{c_{j, 1}, ..., c_{j, l}\}$, i.e., $\boldsymbol{\beta}$ corresponds to the row-wise maxima of $\mathbf{C}$. Finally, feature vectors are mapped onto perceptual quality scores via a support vector regression. Whereas the codebook is optimized unsupervisedly, optimizing the regression requires quality annotations.\\
The computational architecture of CORNIA can also be interpreted as a shallow neural network with a single hidden layer (codebook), a non-linear activation function (soft-encoding function) and a fully-connected output layer (SVR). As such it directly relates to more modern neural network-based visual quality models which are usually much deeper, contain many more parameters and are optimized end-to-end. 

\subsection{Random codebooks}
As a baseline to a learned codebook of image prototypes, we construct a codebook by randomly sampling patches from natural images. 
Patches are individually standardized but no whitening is applied to the codebook. This feature model hence represents (some) natural image statistics but contains no learned parameters.\\
As a second baseline, we construct three codebooks from i.i.d. samples obtained respectively from a Normal$(0,1)$, Laplace$(0,1)$ or Uniform$(0,1)$ distribution. Random codebooks represent neither knowledge of the human visual system nor of natural image statistics.

\section{Experimental setup}
\label{sec:experimental_setup}
To evaluate the performance of different codebook models, we follow the experimental setup of \cite{Ye2012a}.
A test image is converted to grayscale and represented by 10000 randomly sampled 7x7 patches. 
Features are then extracted with one of the codebook models.
In case of CORNIA, we use the original codebook
which contains 10000 codes that were optimized on 7x7 patches extracted from distorted image of the CSIQ database \cite{larson2010most}. 
For our first baseline model, we use randomly sampled 7x7 patches from the reference images of CSIQ to construct a codebook of 10000 patches.
For the second baseline model, random noise codebooks are generated with the respective sampling functions from scipy. All codebooks have the same number of parameters ($49\times10000$).\\
For the support vector regression, we use the NuSVR model as provided in scikit-learn with default
parameters (linear kernel, $\eta=0.5$, $C=1$).
The SVR is optimized on the LIVE database \cite{sheikh2006statistical} which we split by reference image
into a training set comprising the distorted images associated with 23 ($\approx 80\%$) reference images and a test set containing the distorted images of the remaining 6 reference images.
Before training the SVR, features of the training set are individually scaled to the range [-1, 1]. 
The corresponding scaling parameters are also applied to the test data.
To obtain reliable results, we repeat all experiments on 10 random data splits. 
We also evaluate the trained models on the TID2013 \cite{ponomarenko2015image} and CSIQ \cite{larson2010most} databases. 
TID2013 and LIVE share some of their reference images but we ensure that the TID2013 test data contains only images that were not part of the training set. Code for all experiments is available at \href{https://github.com/fraunhoferhhi/CuriouslyEffectiveIQE}{https://github.com/fraunhoferhhi/CuriouslyEffectiveIQE}.

\section{Results}
\label{sec:results}
Average test set performances on the LIVE database are presented in \Tabref{tab:pcc_live} in terms of Pearson and Spearman rank-order correlations with MOS values. Models are trained on the full LIVE database, but we
evaluate them on both the full database and distortion type specific
subsets to assess the generalization across the available distortion types. 
We observe that the performance differs only marginally between all models for both
the full dataset as well as distortion specific subsets. Remarkably, even
codebooks of image patches or noise achieve high correlations, comparable to those achieved with a learned codebook.
This result is surprising as it questions the importance of learning or capturing perceptually relevant features. In the following, we further analyze this finding.

\subsection{Cross database evaluation}
In our first analysis we assess whether our findings generalize beyond the
LIVE dataset and evaluate the trained models on TID2013 and CSIQ. Results in \Tabref{tab:pcc_tid_csiq} show that correlations on 
full databases drop dramatically for both TID2013 and CSIQ. This can be explained by the fact that CSIQ and especially TID2013 contain additional, much more diverse distortion types that were not part of the training data. 
We also note performance declines for evaluations on individual distortion types that are part of the training set. However, except for additive white gaussian noise (awgn) distortions, correlations on distortion types that were part of the training set are still fairly strong for all models. In particular, codebooks from patches or noise perform overall on par with learned codebooks, demonstrating that our finding is not limited to the LIVE database. Interestingly, even though CORNIA and the codebook from patches were assembled from CSIQ, this does not lead to better generalization performance on the CSIQ database.

\begin{table}[t]
\centering
\caption{Average Pearson (pcc) and Spearman rank-order (srocc) correlations across 10 test splits on LIVE \cite{sheikh2006statistical}. ``shared'' refers to the subset of distortions that is shared between LIVE, TID2013 and CSIQ (jpeg, jp2k, gblur, awgn).}
\label{tab:pcc_live}
{\tabcolsep=2pt\def\arraystretch{1.2}
\resizebox{1.01\columnwidth}{!}{%
\begin{tabular}{@{}l|l@{}|cc|cc|cc|cc|cc|cc@{}}
\multicolumn{2}{l|}{\multirow{2}{*}{}}   						& \multicolumn{2}{c|}{full} & \multicolumn{2}{c|}{jpeg} & \multicolumn{2}{c|}{jp2k} & \multicolumn{2}{c|}{gblur} & \multicolumn{2}{c|}{awgn} & \multicolumn{2}{c}{shared}  \\ 
\cline{3-14}
\multicolumn{2}{l|}{}                     						& pcc  & srocc               & pcc  & srocc               & pcc  & srocc               & pcc  & srocc                & pcc  & srocc               & pcc  & srocc            \\ 
\hline
cornia  & \multirow{5}{*}{\rotcell{\phantom{Q}LIVE\phantom{Q}}} & 0.93 & \bf0.93               & 0.93 & 0.91               & \bf0.91 & \bf0.91            & \bf0.95 & \bf0.94             & \bf0.96 & \bf0.95            & 0.92 & \bf0.93  \\ 
\cline{1-1}
patches &                                 						& \bf0.94 & \bf0.93            & \bf0.95 & \bf0.93            & \bf0.91 & \bf0.91            & \bf0.95 & \bf0.94             & 0.95 & 0.93               & \bf0.94 & \bf0.93 \\ 
\cline{1-1}
laplace &                                 						& 0.93 & 0.92               & 0.94 & \bf0.93               & 0.88 & 0.89               & \bf0.95 & \bf0.94             & \bf0.96 & 0.94            & 0.92 & 0.92 \\ 
\cline{1-1}
normal  &                                 						& 0.93 & \bf0.93               & 0.94 & \bf0.93               & 0.9  & \bf0.91               & \bf0.95 & \bf0.94             & \bf0.96 & 0.94            & 0.93 & \bf0.93 \\ 
\cline{1-1}
uniform &                                 						& 0.92 & 0.92               & 0.93 & 0.91               & 0.9  & \bf0.91               & 0.94 & 0.93                & 0.95 & 0.93               & 0.92 & 0.92 \\
\end{tabular}
}
}
\end{table}

\begin{table}
\centering
\caption{Cross-database evaluation on TID2013 \cite{ponomarenko2015image} and CSIQ \cite{larson2010most}. Average Pearson (pcc) and Spearman rank-order (srocc) correlations across 10 models trained on LIVE.}
\label{tab:pcc_tid_csiq}
{\tabcolsep=2pt\def\arraystretch{1.2}
\resizebox{\columnwidth}{!}{%
\begin{tabular}{@{}l|l@{}|cc|cc|cc|cc|cc|cc@{}}
\multicolumn{2}{l|}{  \multirow{2}{*}{}}  & \multicolumn{2}{c|}{full} & \multicolumn{2}{c|}{jpeg} & \multicolumn{2}{c|}{jp2k} & \multicolumn{2}{c|}{gblur} & \multicolumn{2}{c|}{awgn} & \multicolumn{2}{c}{shared}  \\ 
\cline{3-14}
\multicolumn{2}{l|}{}                     & pcc      & srocc            & pcc  & srocc            & pcc    & srocc        & pcc   & srocc        & pcc  & srocc        & pcc  & srocc               \\ 
\hline
cornia&\multirow{5}{*}{\rotcell{TID2013}} & \bf0.55  & 0.47             & 0.85 & 0.79             & 0.87   & \bf0.86         & 0.86  & 0.86         & 0.51 & 0.51         & 0.82    & 0.8             \\ 
\cline{1-1}
patches &                                 & \bf0.55  & \bf0.48             & 0.81 & 0.76             & \bf0.89 & \bf0.86        & 0.89  & 0.9          & 0.43 & 0.45         & 0.82    & 0.79            \\ 
\cline{1-1}
laplace &                                 & 0.52     & 0.44             & \bf0.88 & \bf0.84          & 0.88   & \bf0.86         & \bf0.91  & \bf0.91      & \bf0.78 & \bf0.79      & \bf0.86 & \bf0.85              \\ 
\cline{1-1}
normal  &                                 & 0.49     & 0.44             & 0.85 & 0.8              & 0.81   & 0.8          & 0.86  & 0.87         & 0.74 & 0.76         & 0.8     & 0.8            \\ 
\cline{1-1}
uniform &                                 & 0.49     & 0.44             & 0.85 & 0.82             & 0.85   & 0.85         & 0.86  & 0.86         & 0.72 & 0.74         & 0.82    & 0.81             \\ 
\hline
cornia  & \multirow{5}{*}{\rotcell{CSIQ}} & 0.74     & 0.66             & \bf0.92 & \bf0.89          & \bf0.92 & \bf0.89        & \bf0.93  & \bf0.9       & 0.71 & 0.71         & \bf0.9  & \bf0.88 \\ 
\cline{1-1}
patches &                                 & 0.72     & 0.64             & 0.88 & 0.85             & 0.9    & 0.87         & 0.9   & 0.87         & 0.54 & 0.53         & 0.86    & 0.81 \\ 
\cline{1-1}
laplace &                                 & \bf0.76  & \bf0.68             & \bf0.92 & 0.87          & 0.91   & 0.87         & 0.92  & 0.89         & \bf0.75 & 0.76      & \bf0.9  & 0.87 \\ 
\cline{1-1}
normal  &                                 & 0.75     & 0.66             & \bf0.92 & 0.88          & \bf0.92   & 0.88      & 0.91  & 0.89         & 0.74 & 0.76         & 0.89    & 0.87 \\ 
\cline{1-1}
uniform &                                 & 0.75     & 0.66             & \bf0.92 & 0.88          & \bf0.92   & \bf0.89      & 0.91  & \bf0.9          & \bf0.75 & \bf0.78      & \bf0.9  & \bf0.88
\end{tabular}
}
}
\end{table}

\subsection{How well do the best codes perform?}
\label{subsec:double_descent}
\begin{figure}[t!]
	\begin{minipage}[b]{1.\linewidth}
		\centering
		\subfloat{\includegraphics[width=8.5cm]{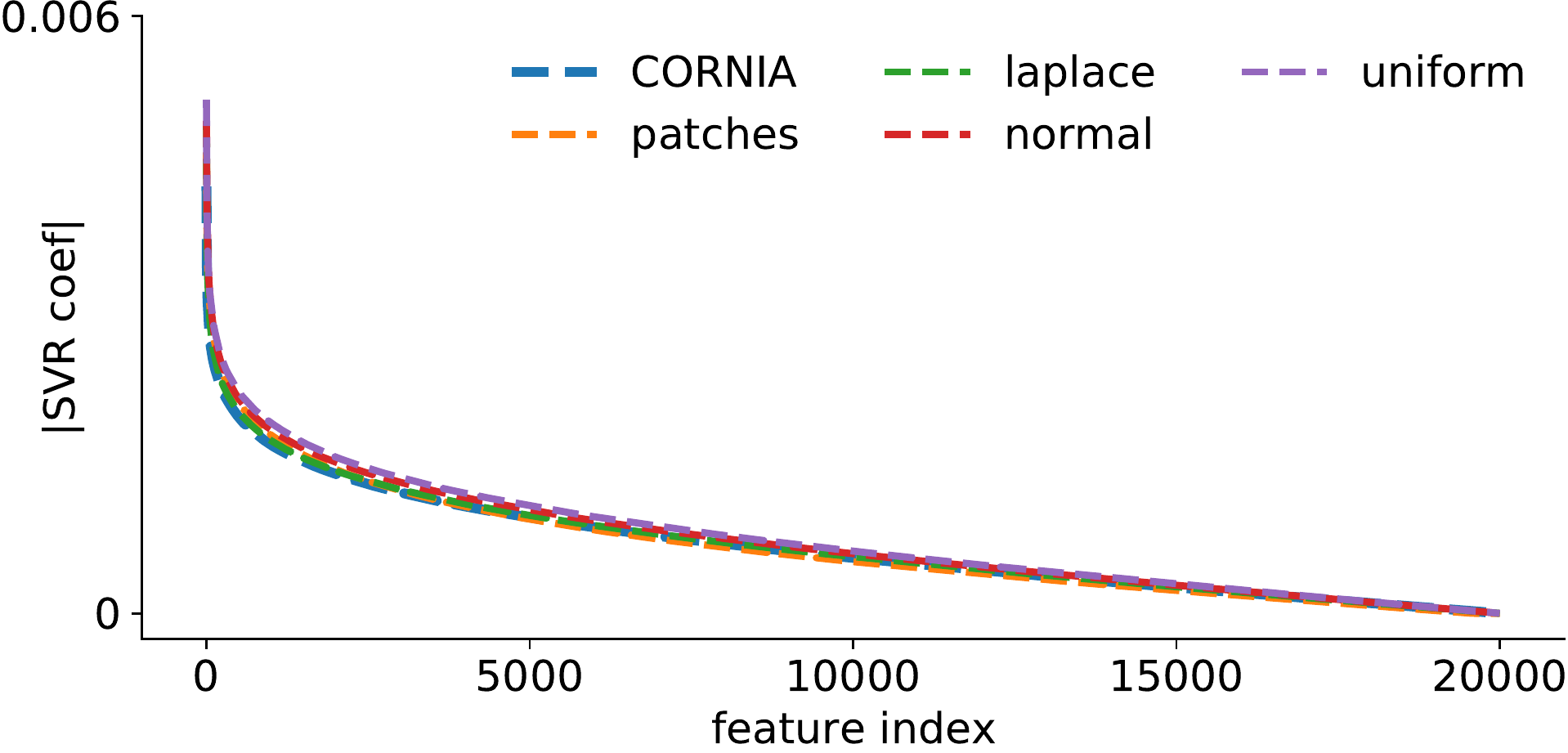}}
	\end{minipage}
	\caption{Absolute SVR coefficient for individual features.}
	\label{fig:svr_coefs}
\end{figure}

For a linear SVR whose input features are all scaled to the same range, the importance of individual features can be assessed by the absolute value of the learned SVR coefficients. As depicted in \Figref{fig:svr_coefs}, there is a relatively small and approximately equal number of highly important features in all models.%
\begin{figure}[t!]
	\begin{minipage}[b]{1.\linewidth}
		\centering
		\subfloat{\includegraphics[width=8.5cm]{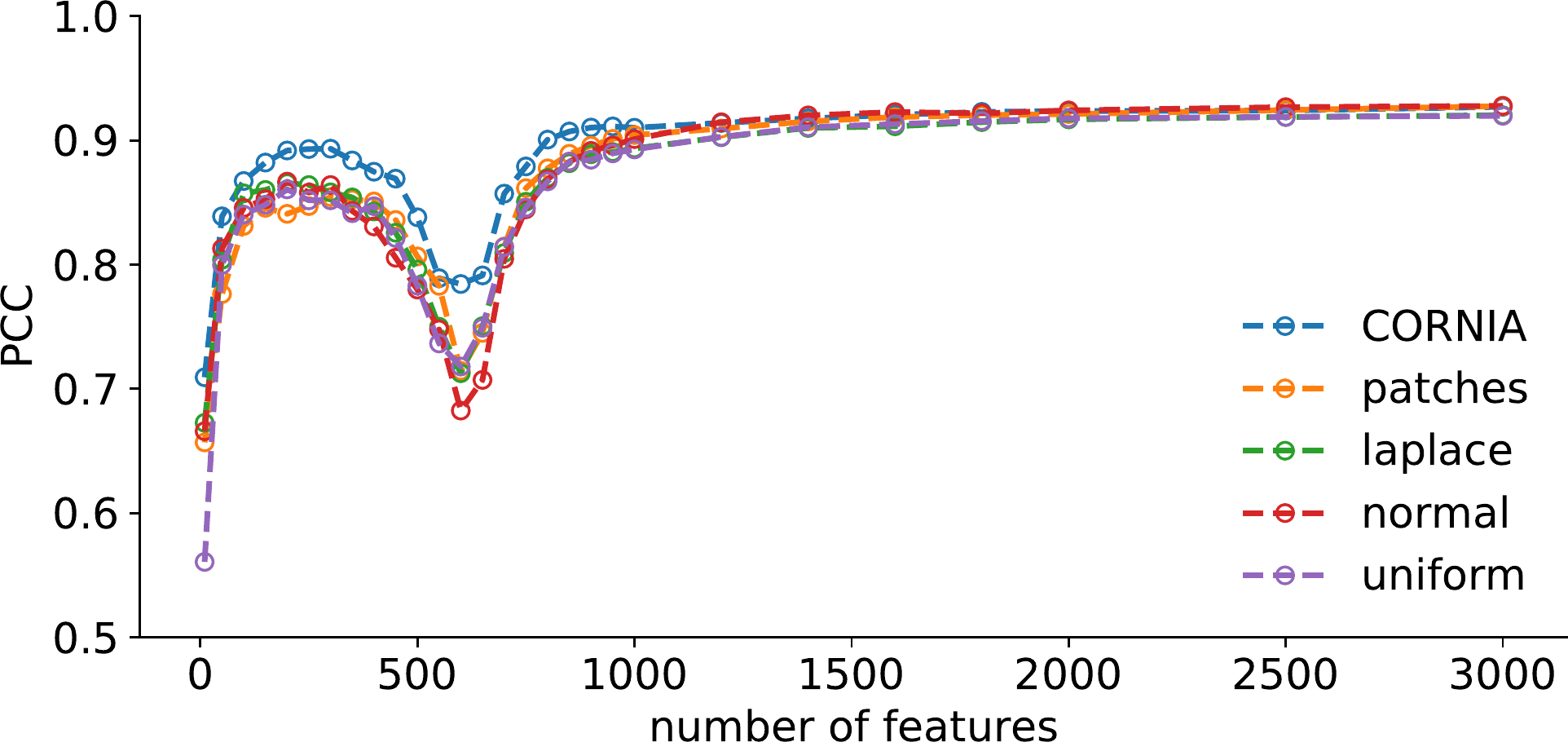}}
	\end{minipage}
	\caption{Average Pearson correlation across 10 random test sets on LIVE as a function of the number of best features.
	}
	\label{fig:pcc_vs_topk}
\end{figure}
Analyzing the SVR coefficients allows us to select the subset of $m$ most important features per model. We vary $m$ to collect subsets of different sizes and train an SVR for each of these using the same hyperparameters as in \Secref{sec:experimental_setup}. We repeat this experiment 10 times on the same random train/test splits of the LIVE database as before.
The average Pearson correlation is depicted in \Figref{fig:pcc_vs_topk} as a function of the number of selected features. For all models the performance follows qualitatively the same curve: initially the correlation increases with the number of features, then descends again, before it finally increases and saturates at a global maximum. This behavior strongly resembles the $\textit{double descent}$ effect that has been described for various classes of models, including linear regression \cite{belkin2019reconciling,mei2019generalization}. In particular, it has been shown that the test set performance of many models as a function of the number of parameters reaches a minimum at the interpolation point, the point at which the number of learned parameters matches the number of training samples ($\approx 605$ in our case, considering only SVR parameters). In addition, these models often reach their peak performance in highly overparameterized regimes. Both of these characteristics are evident in our results for all models.
\begin{figure*}[th!]
	\begin{minipage}[b]{\textwidth}
		\centering
		\vspace{-0.1cm}
		\subfloat{\includegraphics[width=0.6\linewidth]{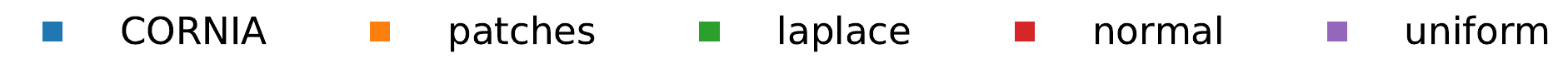}}\\
		\vspace{-0.4cm}
		\setcounter{subfigure}{0}
	    \subfloat[top 300]{\includegraphics[width=0.2\linewidth]{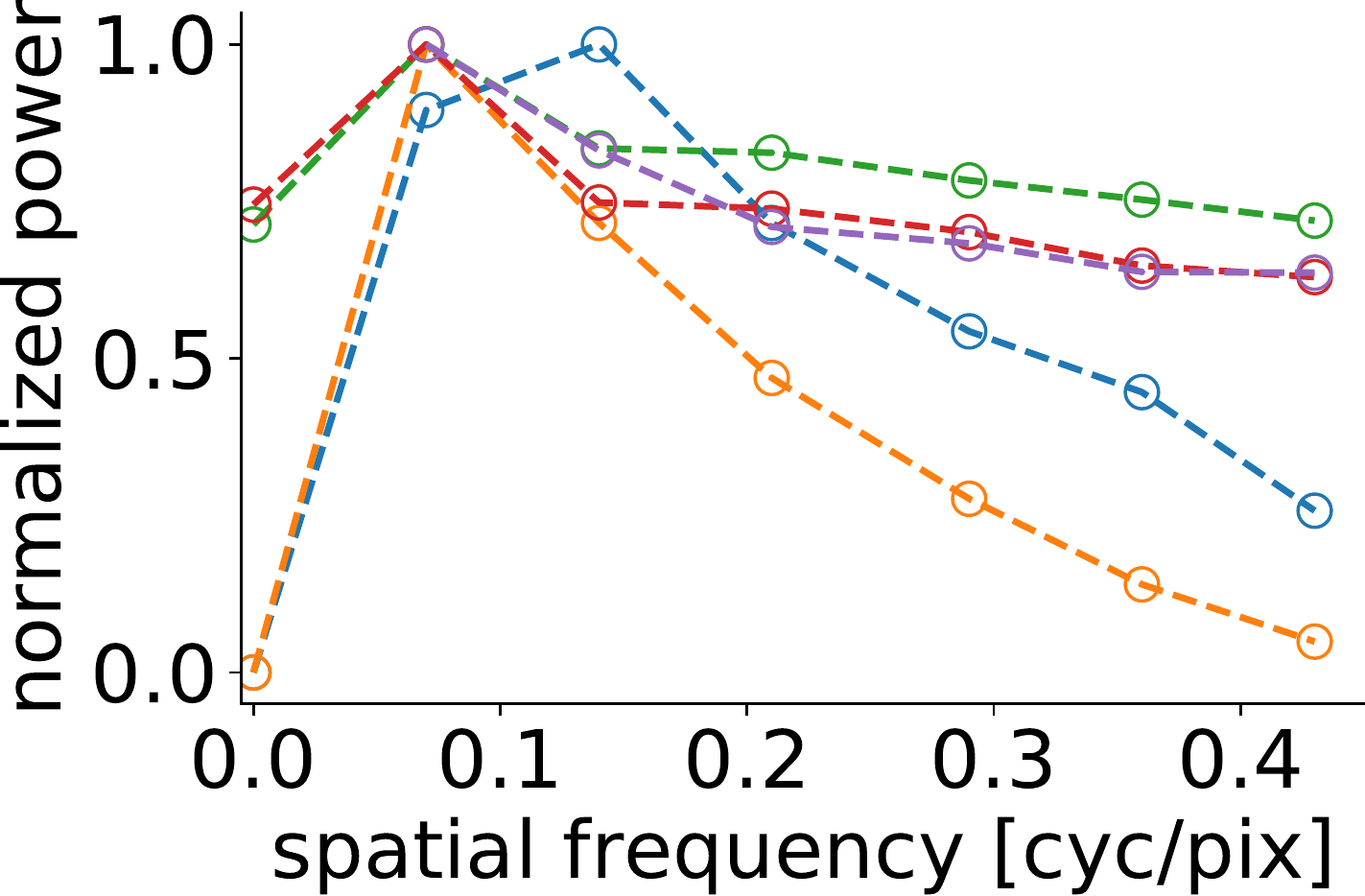}}
	    \subfloat[top 300-600]{\includegraphics[width=0.2\linewidth]{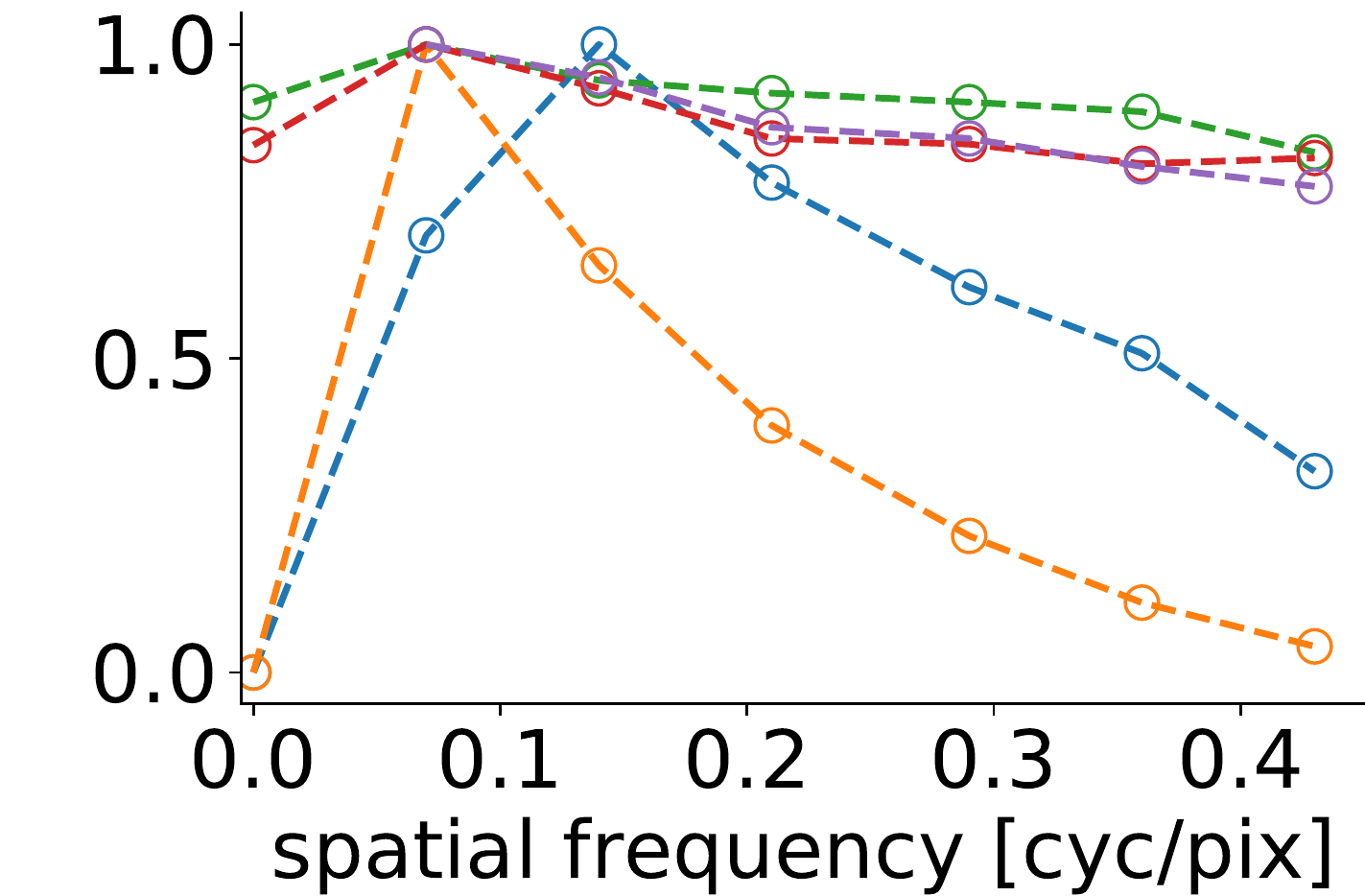}}
	    \subfloat[top 600-900]{\includegraphics[width=0.2\linewidth]{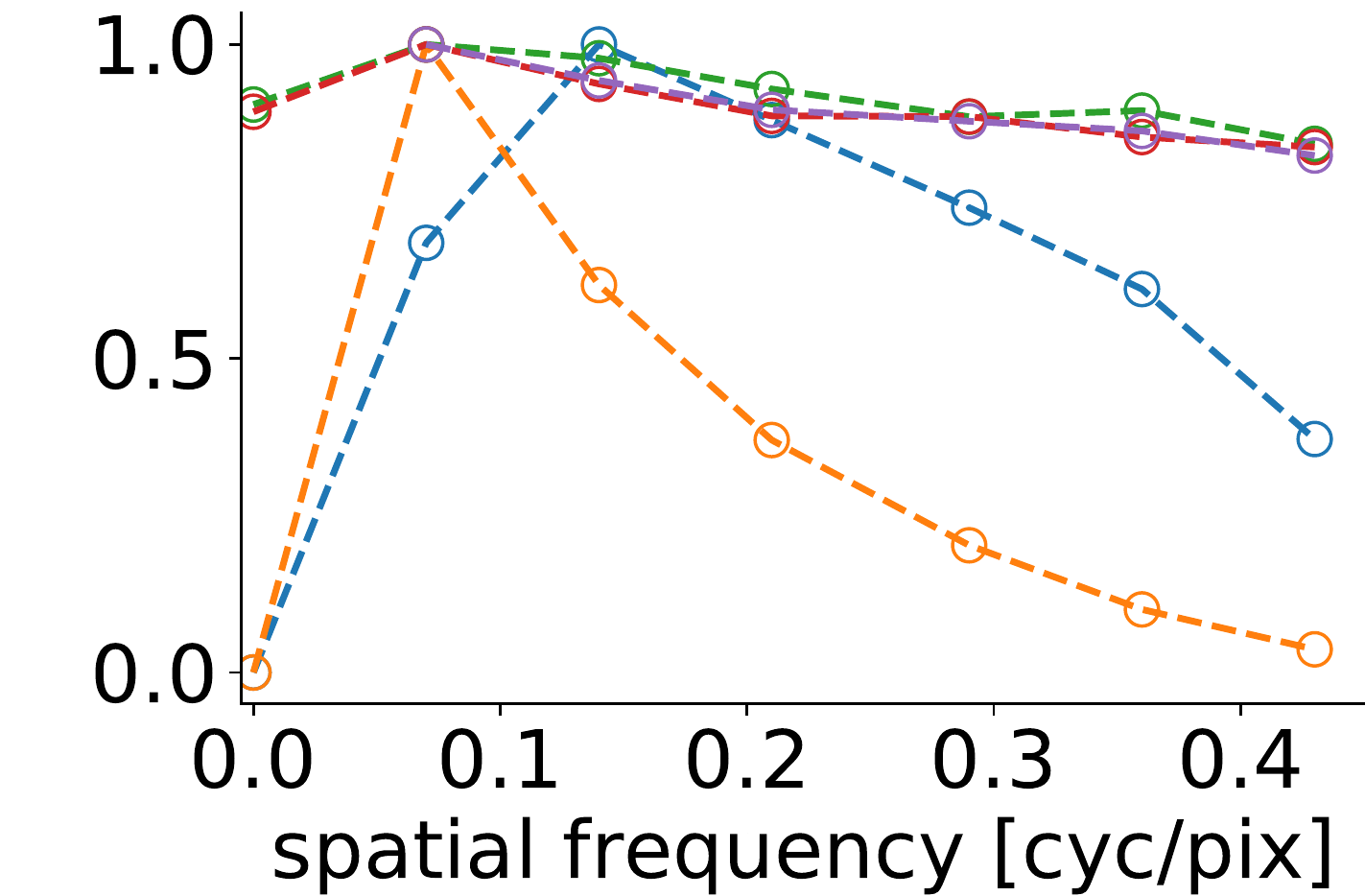}}
	   	\subfloat[top 900-1200]{\includegraphics[width=0.2\linewidth]{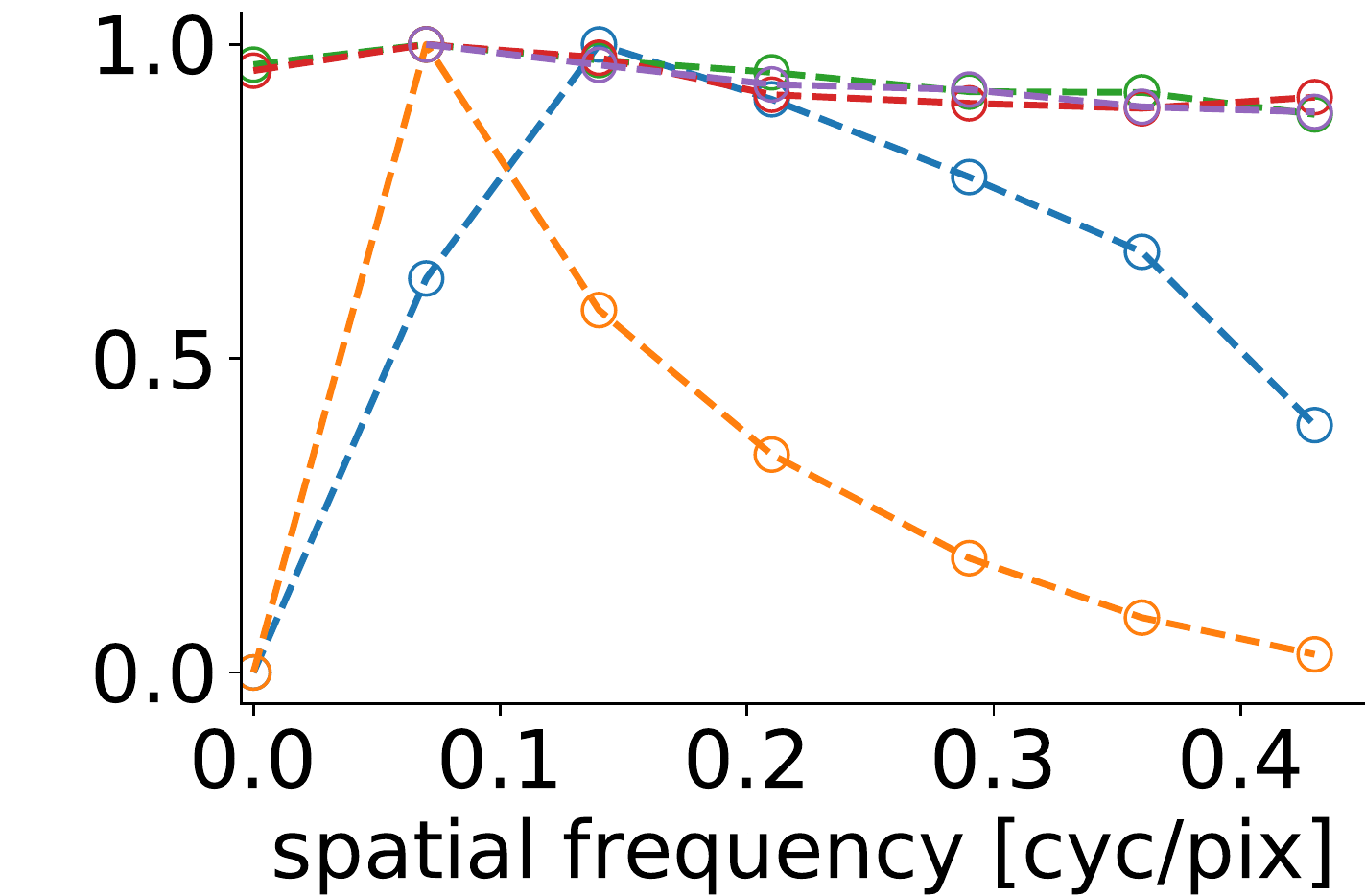}}
	    \subfloat[top 1200-1500]{\includegraphics[width=0.2\linewidth]{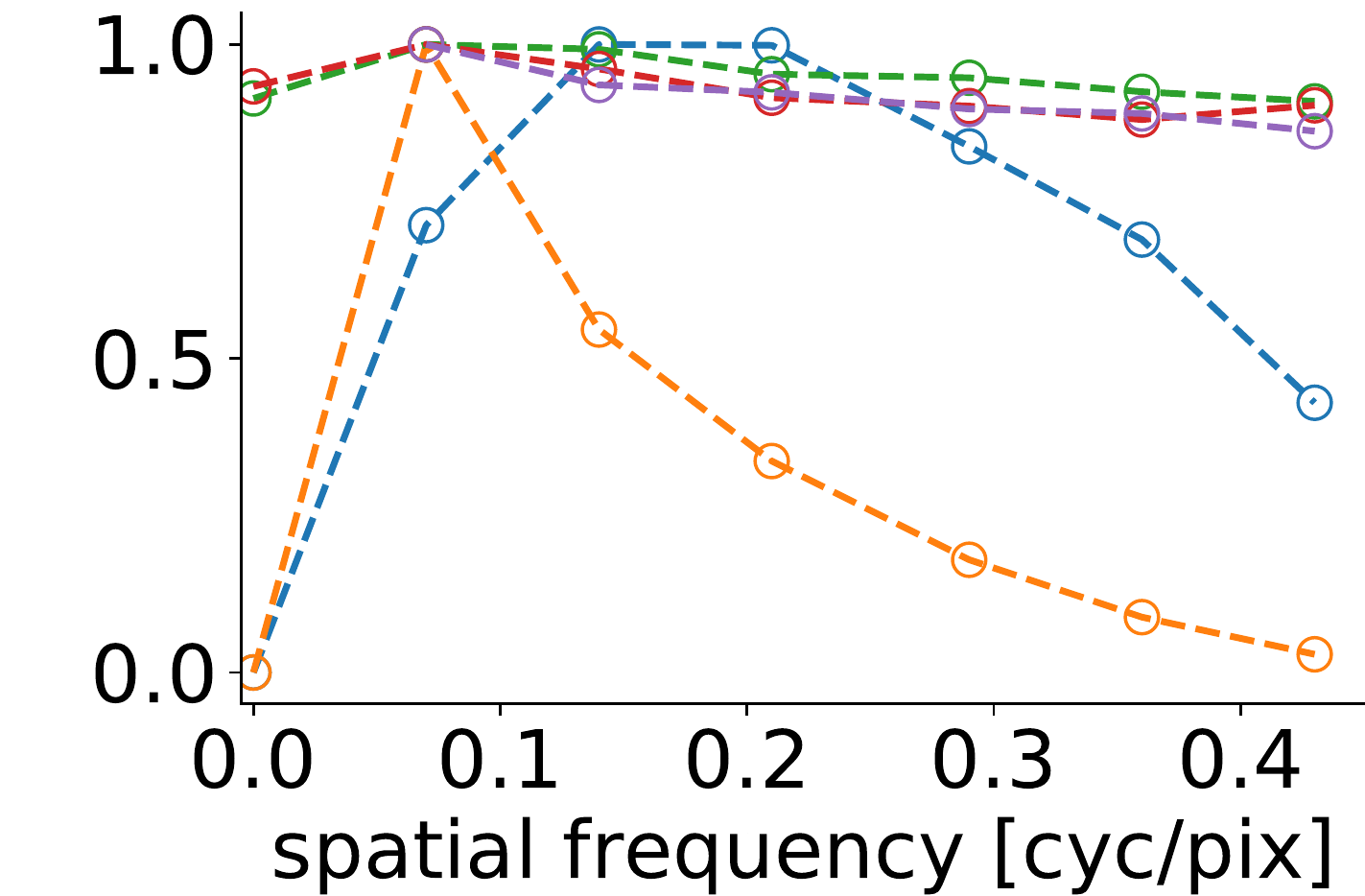}}
	\end{minipage}
	\vspace{-0.2cm}
   	\caption{Normalized power spectra averaged across codes in a subset. The subset in a) holds the 300 most important codes, the subset in b) holds the 301th - 600th most important codes, the last subset e) holds the 1200th - 1500th most important codes.}
   	\label{fig:power_spectra}
\end{figure*}
For small numbers of features, CORNIA consistently outperforms random codebooks by a considerable margin. However, the overall best performance is obtained with large numbers of features with differences between models becoming negligable for 2000 or more features.
Complementary, we can also train models on the 18000 least important features. In this case, the largest deviation from the respective peak Pearson correlations of any model is less than 0.02, demonstrating that even the less important features can lead to accurate predictions.

\subsection{Spectral analysis}
Although the most important codes of CORNIA are more effective than random codes, the overall similarity in performance is still remarkable. 
Did we, by chance, generate feature extractors that capture perceptually-relevant characteristics? 
To assess this hypothesis, we perform a spectral analysis of those codes of a model that generate the most important features according to their regression coefficients. We select the codes corresponding to the 1500 most important features per model and divide them into five mutually exclusive subsets. 
The first subset contains the codes corresponding to the 300 most important features, the last subset corresponds to the codes of the 1200th-1500th most important features. Power spectra, radially-averaged in the frequency domain and averaged across all codes of a subset, are
presented in \Figref{fig:power_spectra}; for a better comparison we normalized
spectra per model.
For the learned codebook of CORNIA, the power spectrum reveals a
distinctive curve on all subsets that resembles the band-pass characteristic of the human contrast sensitivity function \cite{Campbell1968}.
The codebook constructed from random patches exhibits the ${1}/{f^c}$ spectral shape on all subsets well known for natural images \cite{Tolhurst1992}. 
The shapes of the power spectra of random noise codebooks on the other hand vary across the subsets. The spectra for the most important codes show a peak for low (non-dc) frequencies that coincides approximately with the peak in the learned codebook and the codebook of patches. As the codes become less important, the peak levels off and the shape approaches the expected flat white noise spectrum. 
We conclude that some of the random noise feature extractors indeed capture similar features as the learned codebook which partially explains their surprisingly good performance. 
Interestingly, by considering \Figref{fig:pcc_vs_topk} and \Figref{fig:power_spectra} we observe that the similarity between learned codes and random codes decreases in exactly the regime of overparametrization in which all models reach the highest correlations with MOS.

\section{Conclusion}
\label{sec:conclusion}
In this paper we reevaluated the importance of perceptually relevant features for visual image quality prediction. To our surprise, we found that even random feature extractors can lead to high correlations between quality predictions and human quality ratings for a linear regression model. This effect generalizes across distortion types and datasets. In our analyses we have identified two underlying principles for this result. Firstly, by analyzing the power spectra of visual codebooks we found that the most important feature extractors capture similar signal aspects in all models. At first sight it may seem surprising that random noise feature extractors capture perceptually relevant aspects, however, we conjecture that this is a lottery effect \cite{Ramanujan_2020_CVPR} due to the large number of feature extractors (10000) relative to the feature dimension (49). Secondly, we showed that the performance of all models depends critically on the number of features with peak performances being achieved in the highly overparameterized regime. Interestingly, in the overparameterized regime, the similarity between learned and random feature extractors diverges immensely. Taken together, we conclude that visual quality models benefit from feature extractors that capture perceptually relevant aspects, yet, having sufficiently many feature extractors can not only compensate but even outperform a smaller set of individually better feature extractors.\\
The present study is limited to a single computational model. However, as described in \Secref{sec:computational_framework}, this model directly relates to more sophisticated neural networks that constitute the current state of the art in the field. In future work we intend to assess the generalization of our results to state of the art neural networks as well as the generalization from the no-reference to the full reference setting. In addition, we aim to further analyze the capabilities of random feature extractors and are interested in whether these can also be used to optimize image processing systems or merely to monitor visual quality.

\bibliographystyle{IEEEbib}
\bibliography{bwb_icip2021.bib}

\end{document}

%% file: defines.tex
\usepackage[cal=boondox]{mathalfa}

\newcommand{\Figref}[1]{Fig.~\ref{#1}}
\newcommand{\Secref}[1]{Sec.~\ref{#1}}

\newcommand{\Tabref}[1]{Table~\ref{#1}}

%% file: bwb_icip2021.bbl
\begin{thebibliography}{10}

\bibitem{chandler2013seven}
D.~M. Chandler,
\newblock ``Seven challenges in image quality assessment: past, present, and
  future research,''
\newblock {\em International Scholarly Research Notices}, vol. 2013, 2013.

\bibitem{Bosse2019b}
S.~Bosse, M.~Dietzel, S.~Becker, C.~Helmrich, M.~Siekmann, H.~Schwarz,
  D.~Marpe, and T.~Wiegand,
\newblock ``{Neural Network Guided Perceptually Optimized Bit-Allocation for
  Block-Based Image and Video Compression},''
\newblock in {\em IEEE Int. Conf. Image Process.}, 2019, pp. 126--130.

\bibitem{wang2004image}
Z.~Wang, A.~C. Bovik, H.~R. Sheikh, and E.~P. Simoncelli,
\newblock ``Image quality assessment: from error visibility to structural
  similarity,''
\newblock {\em IEEE Trans. Image Process.}, vol. 13, no. 4, pp. 600--612, 2004.

\bibitem{Wang2003}
Z.~Wang, E.~P. Simoncelli, and A.~C. Bovik,
\newblock ``{Multiscale structural similarity for image quality assessment},''
\newblock in {\em IEEE Asilomar Conf. Signals, Syst. Comput.}, 2003, pp.
  1398--1402.

\bibitem{zhang2011fsim}
L.~Zhang, L.~Zhang, X.~Mou, and D.~Zhang,
\newblock ``{FSIM}: A feature similarity index for image quality assessment,''
\newblock {\em IEEE Trans. Image Process.}, vol. 20, no. 8, pp. 2378--2386,
  2011.

\bibitem{Mittal2012}
A.~Mittal, A.~K. Moorthy, and A.~C. Bovik,
\newblock ``{No-reference image quality assessment in the spatial domain},''
\newblock {\em IEEE Trans. Image Process.}, vol. 21, no. 12, pp. 4695--4708,
  2012.

\bibitem{Saad2012}
M.~A. Saad, A.~C. Bovik, and C.~Charrier,
\newblock ``{Blind image quality assessment: A natural scene statistics
  approach in the DCT domain},''
\newblock {\em IEEE Trans. Image Process.}, vol. 21, no. 8, pp. 3339--3352,
  2012.

\bibitem{Ye2012a}
P.~Ye, J.~Kumar, L.~Kang, and D.~Doermann,
\newblock ``{Unsupervised feature learning framework for no-reference image
  quality assessment},''
\newblock in {\em IEEE Conf. Comput. Vis. Pattern Recognit.}, 2012, pp.
  1098--1105.

\bibitem{Zhang2015}
P.~Zhang, W.~Zhou, L.~Wu, and H.~Li,
\newblock ``{SOM: Semantic obviousness metric for image quality assessment},''
\newblock {\em IEEE Conf. Comput. Vis. Pattern Recognit.}, pp. 2394--2402,
  2015.

\bibitem{SPCA}
Z.~Yun, W.~Chao, and M.~Xuanqin,
\newblock ``{SPCA: a no-reference image quality assessment based on the
  statistic property of the PCA on nature images},''
\newblock in {\em Digital Photography IX}. 2013, vol. 8660, pp. 129 -- 136,
  SPIE.

\bibitem{Zhang2018}
R.~Zhang, P.~Isola, A.~A. Efros, E.~Shechtman, and O.~Wang,
\newblock ``{The Unreasonable Effectiveness of Deep Features as a Perceptual
  Metric},''
\newblock {\em IEEE Conf. Comput. Vis. Pattern Recognit.}, pp. 586--595, 2018.

\bibitem{Bhardwaj2020}
S.~Bhardwaj, I.~Fischer, J.~Ball{\'{e}}, and T.~Chinen,
\newblock ``{An Unsupervised Information-Theoretic Perceptual Quality
  Metric},''
\newblock {\em arXiv Prepr. arXiv2006.06752}, 2020.

\bibitem{chetouani2020image}
A.~Chetouani,
\newblock ``Image quality assessment without reference by mixing deep
  learning-based features,''
\newblock in {\em IEEE Int. Conf. Multi. and Expo. (ICME)}, 2020, pp. 1--6.

\bibitem{Bosse2018}
S.~Bosse, D.~Maniry, K.-R. M{\"{u}}ller, T.~Wiegand, and W.~Samek,
\newblock ``{Deep neural networks for no-reference and full-reference image
  quality assessment},''
\newblock {\em IEEE Trans. Image Process.}, vol. 27, no. 1, pp. 206--219, 2018.

\bibitem{ding2020image}
K.~Ding, K.~Ma, S.~Wang, and E.~P. Simoncelli,
\newblock ``Image quality assessment: Unifying structure and texture
  similarity,''
\newblock {\em arXiv preprint arXiv:2004.07728}, 2020.

\bibitem{Prashnani2018}
E.~Prashnani, H.~Cai, Y.~Mostofi, and P.~Sen,
\newblock ``{PieAPP: Perceptual Image-Error Assessment Through Pairwise
  Preference},''
\newblock {\em IEEE Conf. Comput. Vis. Pattern Recognit.}, pp. 1808--1817,
  2018.

\bibitem{Bosse2018distsens}
S.~Bosse, S.~Becker, K.-R. M{\"{u}}ller, W.~Samek, and T.~Wiegand,
\newblock ``Estimation of distortion sensitivity for visual quality prediction
  using a convolutional neural network,''
\newblock {\em Digit. Signal Process.}, vol. 91, pp. 54--65, 2019.

\bibitem{Ramanujan_2020_CVPR}
V.~Ramanujan, M.~Wortsman, A.~Kembhavi, A.~Farhadi, and M.~Rastegari,
\newblock ``{What's Hidden in a Randomly Weighted Neural Network?},''
\newblock in {\em IEEE Conf. Comput. Vis. Pattern Recognit.}, 2020.

\bibitem{belkin2019reconciling}
M.~Belkin, D.~Hsu, S.~Ma, and S.~Mandal,
\newblock ``Reconciling modern machine-learning practice and the classical
  bias--variance trade-off,''
\newblock {\em Proc. Natl. Acad. Sci.}, vol. 116, no. 32, pp. 15849--15854,
  2019.

\bibitem{larson2010most}
E.~C. Larson and D.~M. Chandler,
\newblock ``Most apparent distortion: full-reference image quality assessment
  and the role of strategy,''
\newblock {\em J. electron. imaging}, vol. 19, no. 1, pp. 011006, 2010.

\bibitem{sheikh2006statistical}
H.~R. Sheikh, M.~F. Sabir, and A.~C Bovik,
\newblock ``A statistical evaluation of recent full reference image quality
  assessment algorithms,''
\newblock {\em IEEE Transactions on image processing}, vol. 15, no. 11, pp.
  3440--3451, 2006.

\bibitem{ponomarenko2015image}
N.~Ponomarenko, L.~Jin, O.~Ieremeiev, V.~Lukin, K.~Egiazarian, J.~Astola,
  B.~Vozel, K.m Chehdi, M.~Carli, F.~Battisti, and C.-C.~J. Kuo,
\newblock ``Image database {TID2013}: Peculiarities, results and
  perspectives,''
\newblock {\em Signal process. Image commun.}, vol. 30, pp. 57--77, 2015.

\bibitem{mei2019generalization}
S.~Mei and A.~Montanari,
\newblock ``The generalization error of random features regression: Precise
  asymptotics and double descent curve,''
\newblock {\em arXiv preprint arXiv:1908.05355}, 2019.

\bibitem{Campbell1968}
F.~W. Campbell and J.~G. Robson,
\newblock ``{Application of fourier analysis to the visibility of gratings},''
\newblock {\em J. Physiol.}, vol. 197, no. 3, pp. 551--566, 1968.

\bibitem{Tolhurst1992}
D.~J. Tolhurst, Y.~Tadmor, and Tang Chao,
\newblock ``{Amplitude spectra of natural images},''
\newblock {\em Ophthalmic Physiol. Opt.}, vol. 12, no. 2, pp. 229--232, 1992.

\end{thebibliography}
